\documentclass{article}
\usepackage{arxiv}

\usepackage[utf8]{inputenc} 
\usepackage[T1]{fontenc}    
\usepackage{hyperref}       
\usepackage{url}            
\usepackage{booktabs}       
\usepackage{amsfonts}       
\usepackage{nicefrac}       
\usepackage{microtype}      
\usepackage{lipsum}
\usepackage{graphicx}
\graphicspath{ {./images/} }

\usepackage{bm}
\usepackage{amsmath}
\usepackage{overpic}
\usepackage{algorithm}
\usepackage{algorithmic}
\usepackage{multirow}
\usepackage{diagbox}
\usepackage{graphicx}
\usepackage{subfigure}
\usepackage[justification=centering]{caption}

\title{Online Dynamic Network Embedding}

\author{
 Haiwei~Huang \\
  School of Coumputer and Science Technology\\
  University of Science and Technique of China\\
  Hefei, Anhui 230027\\
  \texttt{hwhuang@mail.ustc.edu.cn} \\
   \And
 Jinlong Li \\
  School of Coumputer and Science Technology\\
  University of Science and Technique of China\\
  Hefei, Anhui 230027\\
  \texttt{jlli@ustc.edu.cn} \\
  \And
 Huimin He \\
  School of Coumputer and Science Technology\\
  University of Science and Technique of China\\
  Hefei, Anhui 230027\\
  \texttt{hehuimin@mail.ustc.edu.cn} \\
  \And
 Huanhuan Chen \\
  School of Coumputer and Science Technology\\
  University of Science and Technique of China\\
  Hefei, Anhui 230027\\
  \texttt{hchen@ustc.edu.cn} \\
}

\begin{document}
\maketitle
\begin{abstract}
Network embedding is a very important method for network data. However, most of the algorithms can only deal with static networks. In this paper, we propose an algorithm Recurrent Neural Network Embedding (RNNE) to deal with dynamic network, which can be  typically divided into two categories: a) topologically evolving graphs whose nodes and edges will increase (decrease) over time; b) temporal graphs whose edges contain time information. In order to handle the changing size of dynamic networks, RNNE adds virtual node, which is not connected to any other nodes, to the networks and replaces it when new node arrives, so that the network size can be unified at different time. On the one hand, RNNE pays attention to the direct links between nodes and the similarity between the neighborhood structures of two nodes, trying to preserve the local and global network structure. On the other hand, RNNE reduces the influence of noise by transferring the previous embedding information. Therefore, RNNE can take into account both static and dynamic characteristics of the network.We evaluate RNNE on five networks and compare with several state-of-the-art algorithms. The results demonstrate that RNNE has advantages over other algorithms in reconstruction, classification and link predictions.
\end{abstract}

\section{Introduction}
\label{sec:Introduction}

Now there is much network structured data like social networks and transportation networks in daily life and research. The real-world networks are often large and complicated so that it's expensive to use them.

Network embedding means learning a low-dimensional representation, e.g., a numerical vector, for
every node in a network. After embedding, other data driven algorithms that need node features as input can be conducted in the low-dimensional space directly. The network embedding is essential in the traditional tasks, such as link predictions, recommendation and classification.  

There are mainly two approaches to conduct network embedding: a) Singular Value Decomposition (SVD)~\cite{SVD} based methods, which is proven to be successful in many important network applications. It decomposes the  adjacency matrix or Laplacian matrix to obtain the node representation. b) Deep learning based methods. Many deep learning based algorithms try to merge structural information to nodes to obtain the low-dimensional representation~\cite{SDNE,deepwalk,line,node2vec}.

The mentioned algorithms of network embedding above are suitable for static networks, in which all the nodes, edges and the features are known and fixed before learning. However, many of the networks are highly dynamic in nature. For example, the social networks, financial transaction networks, telephone call networks, etc., change all the time and remain much information during network evolution.
So when the nodes or edges of the network change, the algorithms need be re-run with the 
whole network data. Usually it will take a long time to learn the embedding again. The online learning of network embedding would be involved temporal analysis, which is similar as dynamic system modelling \cite{li2018symbolic,gong2018sequential,chen2014cognitive}, and its further analysis and work \cite{ChenTRY14,chen2013model,gong2016model,chen2015model}.

Most of the dynamic network embedding algorithms are based on the static network algorithms. They will more or less encounter the following challenges: 
\begin{itemize}
\item \textbf{Network structure preservation}: some algorithms learn representation of new nodes by performing information propagation~\cite{propagation}, or optimizing a loss that encourages smooth changes between linked nodes~\cite{harmonic, non-parametric}. There are also methods that aim to learn a mapping from node features to representations, by imposing a manifold regularizer derived from the graph~\cite{Manifold}. But these methods do not preserve intricate network properties when inferring representation of new nodes. 
\item \textbf{Growing graphs}: Structural Deep Network Embedding (SDNE)~\cite{SDNE} method and Deeply Transformed High-order Laplacian Gaussian Process (DepthLGP)~\cite{DepthLGP} both use a deep neural network to learn representations with considering the network structure. But SDNE could not handle nodes change and DepthLGP could not handle edges change. The SVD based algorithms could not handle growing graphs either. Incremental SVD methods~\cite{fastSVD1, fastSVD2} are proposed to update previous SVD results to incorporate the changes without restarting the algorithm. But it can only deal with edges change and when errors cumulate, it still need to re-run SVD to correct the errors. 
\item \textbf{Information of evolving graphs}: Dynamic Graph Embedding Model (DynGem)~\cite{dyngem} uses a dynamically expanding deep autoencoder to keep network structure and deal with growing graphs. However,  it only trains the current network on the basis of the old parameters and abandons the information contained in the network during the evolution.
\end{itemize}

\begin{figure*}[htbp]
\centering
\subfigure[RNNE cell]{
\begin{minipage}[t]{0.3\linewidth}
\centering
\includegraphics[width=\linewidth]{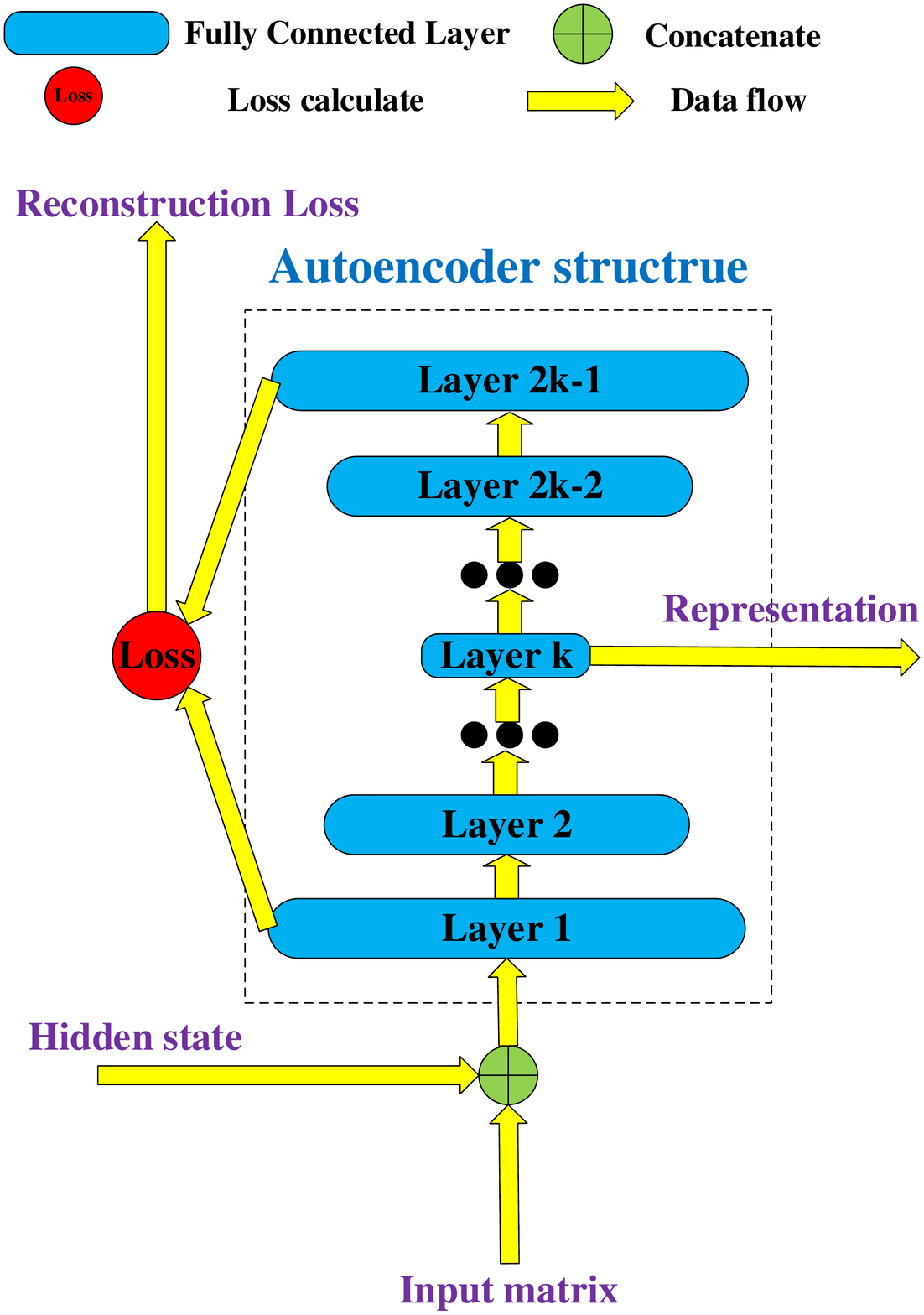}
\end{minipage}
}
\subfigure[RNNE structure]{
\begin{minipage}[t]{0.66\linewidth}
\centering
\includegraphics[width=\linewidth]{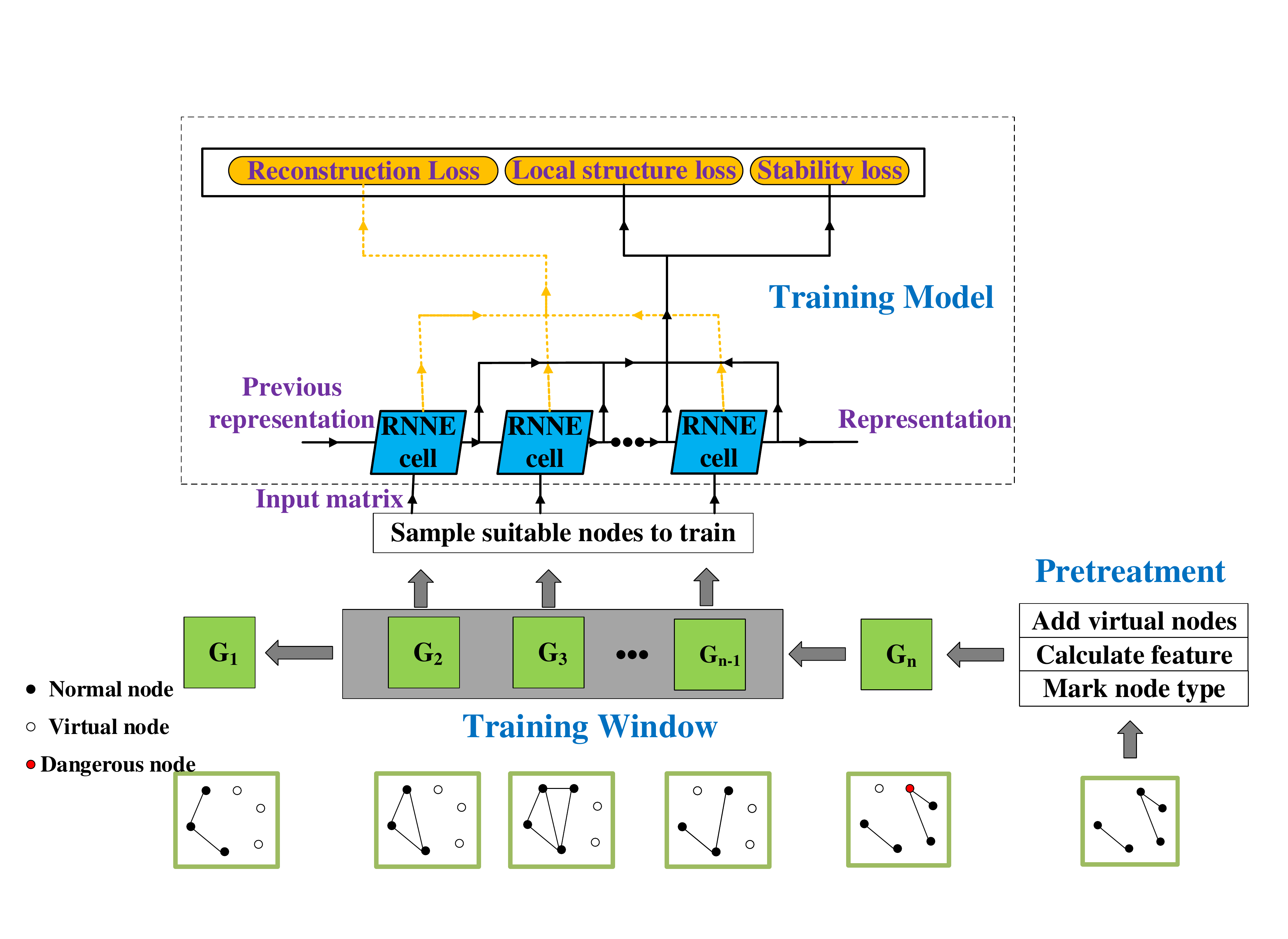}
\end{minipage}
}
\caption{\textbf{(a) is the structure of RNNE cell that is used in (b). (b) is the components and processes of RNNE model, all of the RNNE cells share the same parameters. \bm{$G_{1},...,{G_{n}}$} is the series of networks.}}
\label{fig:RNNE}
\end{figure*}

To improve the embedding of dynamic network, we propose Recurrent Neural Network Embedding (RNNE), a neural network model,  which is shown in Figure \ref{fig:RNNE}. In response to the three challenges mentioned above, RNNE has adopted the following approaches in the three main parts of the model (Pretreatment, Training Window and Training Model):
\begin{itemize}
\item \textbf{Network structure preservation}: RNNE calculates the node features from multi-step probability transition matrices in Pretreatment, trying to preserve the structural characteristics of larger neighborhoods of each node than only using the adjacency matrix. And in Training Model,  the loss function will consider the first-order proximity, high-order proximity\footnote{The first-order proximity is determined by if there is a link between two nodes, and the high-order proximity means the similarity between the neiborhood structure of two nodes.}together.
\item \textbf{Growing graphs}: RNNE will first put some virtual nodes to the network. When new nodes arrive,  RNNE will replace the virtual nodes with new nodes in Pretreatment. Similarly, if a node is deleted, RNNE will replace it with a virtual node.
\item \textbf{Information of evolving graphs}: The overall structure of Training Model is a RNN model. The previous node representations are inputted as hidden state to the RNNE cell, so more information of evolving graphs can be used during embedding. Considering that the representations of one node at different time should be closed if the node's characters don't change, RNNE also adds a corresponding part to the loss function to maintain the stability\footnote{Stability means reducing the effects of noise from network fluctuation over time.}of embedding.
\end{itemize}

The main contributions of this paper are listed as follows:
\begin{itemize}
\item RNNE considers the first-order proximity and high-order proximity during training, so it can preserve the original network structure.
\item With virtual nodes, RNNE can unify the sizes of networks at different time and easily extract the changing part of the network. 
\item RNNE takes graph sequences as input and can integrate information of evolving graphs when embedding. It is helpful to mitigate the effects of network fluctuation over time. 
\end{itemize}

This paper is organized as follows. Section~\ref{sec:RNNE} introduces and explains the RNNE model in detail. And Section~\ref{sec:experiments} introduces the experiments and datasets. The experimental results and analysis are presented in Section~\ref{sec:result}.  Finally, Section~\ref{sec:Conclusion} concludes this paper.

\section{The RNNE model}
\label{sec:RNNE}

\subsection{Problem Definition}
\label{sec:problem}
Given a dynamic network $G$ whose nodes and edges may change when time goes on and then 
given a series of graph $G_{1}(V_{1}, E_{1})$, $G_{2}(V_{2}, E_{2})$, ..., $G_{t}(V_{t}, E_{t})$ where $G_{i}(V_{i}, E_{i})$ is the state of $G(V, E)$ in a series of time, for each node $v$ of $G_{t}$, learn $f:V_{t}\rightarrow R^{K}$, where $K$ is a positive integer given in advance.

\subsection{Model Description}
\label{sec:model}
First RNNE assumes that the network series are stable. It means that there won't be too many nodes changing at the same time, 
and the increase in weight of edges is nearly linear. Second, the size of the model is limited, so RNNE also assume that the network 
will not become too large with time goes on.

RNNE will not process the whole series at the same time, because the old network maybe invalid and too long series will take 
a lot of time. RNNE maintain a fixed length window to get the networks to train and then a concept drift checking part will exclude 
the nodes whose property maybe change. 

For the node in dynamic network, we can't represent it only with a state of a moment. So RNNE not only use the current state and also 
consider the previous state when learning the embedding. Learn from recurrent neural networks (RNN)~\cite{RNN}, RNNE use a hidden state to represent the 
previous state of the node.

In general, RNNE will first choose suitable nodes, then use hidden state and node feature as input to minimize the loss of node 
proximity in neighboring time points. The entire process will be explained in detail in the following subsections.

\subsection{Pretreatment}
\label{sec:pretreatment}
Each node has an attribute named ``state''. At the beginning all the nodes' ``state'' is ``normal'' which means it is only a normal node:
\begin{center}
$v_{ik}$[``state'' ] = ``normal'' ,\\
where $v_{ik}$ is the $i$-th node in $G_{k}$ 
\end{center}
Then in order to keep the size of input, we define a type of node named virtual node. It is not connected to any other node. And 
\begin{center}
$v$[``state'' ] = ``virtual'' ,\\
if $v$ is a virtual node
\end{center}
If the number of nodes $|V_{k}|$ in $G_{k}$ doesn't reach the limit of the model which is $N$, then we add virtual nodes into $G_{k}$ until 
$|V_{k}|$ = $N$.

Before start training, we should add training networks to the training window one by one. When the new network arrives, if it's the first type dynamic network, it can be put into training window directly. Otherwise, RNNE will put the subgraph of the increased part than the last network. If the training window 
is full, RNNE removes the earliest one from training window and then check every node in the window to keep out dangerous nodes when training. 

Assume that the window size is 5 and there are 4 networks $G_{a}$, $G_{a+1}$, $G_{a+2}$, $G_{a+3}$ in the window, then $G_{a+4}$ arrive. 
For every node $v_{i, a+4}$ in $G_{a+4}$, if the state of $v_{i, a+4}$ and $v_{i, a+3}$ are both not ``virtual'', we calculate $||v_{i, a+4}-v_{i, a+3}||^{2}$ 
using the row of their adjacency matrix. Most of the time, $v_{i, a+4}$ and $v_{i, a+3}$ are the same node at different time. At last, we use Grubbs test~\cite{grubuustest1, grubuustest2} 
to find the dangerous nodes whose property most likely change:
\begin{center}
$v_{i, a+4}$[``state'' ] = ``dangerous'' ,\\
if $v_{i, a+4}$ is a dangerous node
\end{center}
The process described above is present in Algorithm~\ref{alg:pretreatment}.
 
In order to keep the high-order proximity, we calculate the node feature as below, assume that $M_{k}\in R^{N\times N}$ is the adjacency matrix of $G_{k}$:\\
First we define a function $normalized$:
\begin{equation}
normalized(A) = \frac{A}{max(A, axis = 1)}
\end{equation}
each element in $A$ will be divided by the largest element in the same row.\\
The feature matrix $X_{k}$ is calculated as follow:
\begin{equation}
\label{eq:feature}
\begin{split}
&U_{k} = normalized(M_{k})\\
&V_{k} = normalized(M_{k}^{2})\\
&W_{k} = normalized(M_{k}^{3})\\
&X_{k} = normalized(\frac{U_{k}}{2}+\frac{V_{k}}{3}+\frac{W_{k}}{6})\\
\end{split}
\end{equation}

\begin{algorithm}[htb] 
    \caption{mark state of new network} 
    \label{alg:pretreatment} 
    \begin{algorithmic}[1] 
    \REQUIRE new network $G(V, E)$ with adjacency matrix $M$, the last network $G^{'}(V^{'}, E^{'})$ with adjacency matrix $M^{'}$, the size limit $N$, the significance level $\alpha$
    \ENSURE mark the state of $v\in V$
    \STATE Add virtual node $V_{v}$ into $G$ until $|V|=N$
    \FOR{each $v$ in $V_{v}$}
    \STATE $v$[``state'']=``virtual''
    \ENDFOR
    \STATE $D = (M-M^{'})^{\circ 2}\times \mathbf{1}_{N\times 1}$ 
    \STATE use Grubbs test on D with ignoring the virtual node data to find target node set $V_{d}$ with significance level $\alpha$
    \FOR{each $v$ in $V_{d}$}
    \STATE $v$[``state'']=``dangerous''
    \ENDFOR
    \FOR{each $v$ in $V-V_{v}-V_{d}$}
    \STATE $v$[``state'']=``normal''
    \ENDFOR
\end{algorithmic}
\end{algorithm}

\subsection{Training}
\label{sec:training}
RNNE model is shown in Figure~\ref{fig:RNNE}. Suppose there are $n$ networks $G_{a+1}$, $G_{a+2}$, ......, $G_{a+n}$ in the window, the symbol is explained in Table~\ref{tab:symbol}.
\begin{table}[h]
\caption{\textbf{Symbol Explanation}}
\label{tab:symbol}
\centering
\begin{tabular}{|c|c|}
\hline
Symbol&Definition\\
\hline
$v_{i}, v_{ik}$&$v_{ik}$ are the same node $v_{i}$ on different time point $k$\\
\hline
$N$&the node size of $G_{k}$ after adding virtual nodes, $j = 1, 2, ..., t$\\
\hline
$d$&the embedding size\\
\hline
$b$&the batch size of training\\
\hline
$x_{ik}$&the feature of the $i$-th node in $G_{k}$ which is calculated as Eq.\ref{eq:feature}, $x_{ik}\in R^{N}$\\
\hline
$\hat{x}_{ik}$&the reconstructed data of $v_{ik}$,$\hat{x}_{ik}\in R^{N+d}$\\
\hline
$M_{k}$&the adjacency matrix for the $G_{k}$, $M_{k}\in R^{N\times N}$\\
\hline
$h_{ik}$&the hidden state of $v_{ik}$, $h_{ik}\in R^{d}$\\
\hline
$y_{ik}$&the representation of $v_{ik}$, $y_{ik}\in R^{d}$\\
\hline
$s_{ik}$&the state of  $v_{ik}, s_{ik}\in \{$``normal', ``virtual'', ``dangerous''$\}$\\
\hline
\end{tabular}
\end{table}

First RNNE sample a batch from nodes, a node $v_{i}$ can be chosen only when $s_{ik} = $``normal'' $(k = a+1, a+2,..., a+n)$. Assume that node index $u_{1}, u_{2},...,u_{b}$ are selected, 
then we get the input matrix series $X_{k} = \{x_{u_{i}, k}\}_{i=1}^{b}(k = a+1, a+2,..., a+n)$.

The RNN cell of our model has an encoder-decoder structure. The encoder $E$ consists of multiple non-linear functions that map the input data to the representation space. The decoder $D$ also consists of
multiple non-linear functions mapping the representations in representation space to reconstruction space. Given the input $x_{u_{i}, k}$ ,the calculation is shown as follows:
\begin{equation}
\label{eq:xyh}
\begin{split}
&y_{u_{i}, k} = E([h_{u_{i}, k-1},x_{u_{i}, k}])\\
&\hat{x}_{u_{i}, k} = D(y_{u_{i}, k})\\
&h_{u_{i}, k} = y_{u_{i}, k}
\end{split}
\end{equation}
The goal of the autoencoder is to minimize the reconstruction error of the output and the input. The loss is calculated as below:
\begin{equation}
L_{k} = \sum_{i=1}^{b}||\hat{x}_{u_{i}, k}-[h_{u_{i}, k-1},x_{u_{i}, k}]||^{2}
\end{equation}
As \cite{hashing} mentioned, although minimizing the reconstruction loss does not explicitly preserve the similarity between samples, the reconstruction criterion can smoothly capture the data manifolds and
thus preserve the similarity between samples. It means that if the input is similar then the output will likely similar. In other words, if the features of two nodes are similar, the embedding of the nodes are similar. Simultaneously, with reconstructing the node feature from the embedding, we can possibly make sure that the embedding vector contains enough information to represent the node.

Considering that there are a lot of zero elements in $M_{k}$ and $x_{u_{i}, k}$, but in fact we are more concerned about the non-zero part in them. Learn from SDNE~\cite{SDNE}, when calculate the reconstruction error, we will add different weight in zero and non-zero element. The new loss function is shown as below:
\begin{equation}
\label{eq:second-order loss}
L_{2,k} = \sum_{i=1}^{b}||(\hat{x}_{u_{i}, k}-[h_{u_{i}, k-1},x_{u_{i}, k}])\odot W_{u_{i}, k})||^{2}
\end{equation}
where $\odot$ means the Hadamard product, $W_{u_{i}, k}=\{w_{u_{i},j, k}\}_{j=1}^{N+d}$. If $m_{u_{i},j,k}\in M_{k} = 0$, $w_{u_{i},j, k} = 1$, else $w_{u_{i},j, k} = \beta > 1$. Using this loss function, the nodes who have similar neighborhood structure will be mapped closely. It means that our model can keep the global network structure by keeping high-order proximity between nodes.

In addition to consider the neighborhood structure of different nodes, we should also pay attention to  the local structure which means the direct link in nodes. We use the first-order proximity to measure the local structure of network. The loss function is shown as below:
\begin{equation}
L_{1,k} = \sum_{i,j=1}^{b}m_{u_{i},u_{j},k}||y_{u_{i},k}-y_{u_{j},k}||^{2}
\end{equation}
if $m_{u_{i},u_{j},k} > 0$, there is a direct link in node $v_{u_{i},k}$ and $v_{u_{j},k}$, we hope them can be mapped near  in the embedding space.

In the above we only considered one network in the series of networks, though using the hidden state to transfer information between them. When we sample the training nodes, the states of them all are ``normal'', so the representation of one node in different time should be close as far as possible without the influence of noise. The loss function of this part is shown as follows:
\begin{equation}
\label{eq:stablity loss}
\begin{split}
&average_{u_{i}} = \frac{\sum_{k=a+1}^{a+n}y_{u_{i}, k}}{n}\\
&L_{time, u_{i}} = \sum_{k=a+1}^{a+n}||y_{u_{i}, k}-average_{u_{i}}||^{2}
\end{split}
\end{equation}

In summary, to keep the first-order proximity and second-order proximity and the stability in time series, we combine Eq.\ref{eq:second-order loss}$\sim$\ref{eq:stablity loss} and get the integrated final loss function:
\begin{equation}
\label{eq:loss}
\begin{split}
L_{total} &= \sum_{k=a+1}^{a+n}(\alpha L_{1,k}+L_{2,k})+\gamma \sum_{i=1}^{b}L_{time,u_{i}}\\
&=\sum_{k=a+1}^{a+n}(\alpha \sum_{i,j=1}^{b}m_{u_{i},u_{j},k}||y_{u_{i},k}-y_{u_{j},k}||^{2}\\
&+\sum_{i=1}^{b}||(\hat{x}_{u_{i}, k}-[h_{u_{i}, k-1},x_{u_{i}, k}])\odot W_{u_{i}, k})||^{2}\\
&+\gamma\sum_{i=1}^{b}||y_{u_{i}, k}-\frac{\sum_{k=a+1}^{a+n}y_{u_{i}, k}}{n}||^{2})
\end{split}
\end{equation}
$\alpha$, $\gamma$ and $\beta$ in $W_{u_{i}, k}$ are the parts of the hyper parameter of the model.

At last the model parameters $\theta$ can be adjusted by:
\begin{equation}
\label{eq:update}
\theta = \theta - \frac{\partial L_{total}}{\partial \theta}\eta
\end{equation}
where $\eta$ is the learning rate of the model.

The whole training process can be seen in Algorithm~\ref{alg:train}.

\begin{algorithm}[htb] 
    \caption{Train RNNE model} 
    \label{alg:train} 
    \begin{algorithmic}[1] 
    \REQUIRE the network list $[G_{k}]_{k=a+1}^{a+n}$ with their  adjacency matrix $[M_{k}]_{k=a+1}^{a+n}$ and feature matrix $[X_{k}]_{k=a+1}^{a+n}$ in trainning window, the hidden state $H_{a}$ of $G_{a}$
    \ENSURE network embedding $[Y_{k}]_{k=a+1}^{a+n}$ and hidden state $[H_{k}]_{k=a+1}^{a+n}$
    \IF {it is the first time to train}
    \STATE initialize the parameter $\theta$
    \ENDIF
    \REPEAT
    \STATE sample a minibatch of nodes $V$ satisfied that : \\for $v$ in $V$, $v_{k}$[``state'']=``normal''
    \STATE get the slice of $V$'s part in $M_{k}$, $X_{k}$, $H_{a}$
    \STATE calculate $L_{total}$ using Eq.\ref{eq:xyh} and Eq.\ref{eq:loss}
    \STATE update parameter $\theta$ with Eq.\ref{eq:update}
    \UNTIL{converge}
    \STATE using Eq.\ref{eq:xyh} to get $[Y_{k}]$ and $[H_{k}]$
\end{algorithmic}
\end{algorithm}

\subsection{Analysis and Discussions}
\label{sec:Analysis and Discussions}
In this section, some analysis and discussions of RNNE are presented. 

RNNE assumes a limit $N$ of node size in each snapshot of network. Usually the node size is less than $N$. If the node size becomes larger than $N$ during network evolution, we can expand the layer size of RNNE cell with ramaining the old parameters, which is learned from DynGem~\cite{dyngem}.

The training complexity of RNNE in one  iteration is $O(nb^{2}(N+d)D)$, where n is the size of trainning window, $b$ is the batch size, $N$ is the limit of node size in network, $d$ is the embedding size, and $D$ is the maximum size of the hidden layer. Usually $n$, $b$ and $d$ are constants given in advance. $N$ is linear to the true node size of network. $D$ is related to the embedding size but not related to the node size. So the training complexity of RNNE in one  iteration is $O(N)$ and linear to the node size of network.
\section{Experiments}
\label{sec:experiments}
In this section, we introduce the methods and datasets which are used to evaluate the RNNE algorithm.

\subsection{Dataset}
\label{sec:dataset}
We use static networks and dynamic networks evaluate the RNNE algorithm. Some of the datasets don not have label information, so they won't be used to do the classification experiment. For static networks, we randomly change some of the nodes and edges to generate a series of networks. All of the dataset's length are 14.
\begin{table}[h]
\caption{\textbf{dataset information}}
\centering
\begin{tabular}{|c|c|c|}
\hline
dataset&nodes&edges\\
\hline 
Wiki&2405-2724&17981-27754\\
\hline
email-Eu-core&1005-1242&25571-43249\\
\hline
blogCatalog&10312-10651&333983-624250\\
\hline
CA-CondMat&23133-23252&93468-176492\\
\hline
CA-HepPh&12008-12337&118505-209384\\
\hline
\end{tabular}
\end{table}
\begin{itemize}
\item Wiki : It is a reference network in wiki and each node has a label. There are totally 17 categories in this dataset.
\item blogCatalog~\cite{blog}, email-Eu-core~\cite{CA} : They are social networks of people. There are 39 categories in blogCatalog and 42 in email-Eu-core.
\item CA-CondMat, CA-HepPh~\cite{CA}: They are collaboration network of Arxiv. These two datasets are only used for the reconstruction and link prediction because we have no label information for of the nodes.
\end{itemize}

\subsection{Baselines and Parameters}
\label{sec:baselines}
We use following algorithms as the baselines of the experiments. For the static network embedding algorithms, we will apply them to each snapshot of the dynamic network.
\begin{itemize}
\item SDNE~\cite{SDNE} : It also uses an autoencoder structure, and learns embedding with minimizing the loss of first-order and second-order proximity. \\
$encoder\_layer\_list = [1000, 128]$, $\alpha = 10^{-6}$, $\beta = 5$, $nu1 = 10^{-5}$, $nu2 = 10^{-4}$.
\item Line~\cite{line} : It doesn't define a function to calculate the network embedding but learning a map of node to embedding directly. It's loss fuction also consider the first-order and second-order proximity.
\item GrapRep~\cite{grarep} : It considers high-order proximity and use SVD to get network embedding.\\
$Kstep = 4$.
\item Hope~\cite{hope} : It constructs an asymmetric relation matrix from the adjacency matrix and then use JDGSVD~\cite{jdgsvd} to get the low-dimensional representation.
\end{itemize}
For RNNE, the layer size\footnote{Layer size is a list of numbers of the neuron at each encoder layer, the last number is the dimension of the output node representation.} of autoencoder is different in each dataset. It is shown in table~\ref{tab:size}.
\begin{table}[h]
\centering
\caption{\textbf{autoencoder size}}
\label{tab:size}
\begin{tabular}{|c|c|}
\hline
dataSet&layer size\\
\hline
Wiki&5128-200-128\\
\hline
email-Eu-core&2128-128\\
\hline
blogCatalog&15128-1000-128\\
\hline
CA-CondMat&25128-2500-1000-128\\
\hline
CA-HepPh&15128-1500-128\\
\hline
\end{tabular}
\end{table}\\
The  hyper-parameters of $\alpha$, $\beta$, $\gamma$ are adjusted by grid search : $\alpha \in [10^{-6}, 1]$, $\beta \in [1, 10]$, $\gamma \in [0, 10]$

\subsection{Evaluation Metrics}
\label{sec:metrics}
In our experiments, we test three tasks of reconstruction, classification and link prediction. 

In reconstruction and link predictions, we use $precision@k$ whose definition is shown below to measure the performance of algorithms.\\
For $G(V, E)$:
\begin{equation*}
precision@k = \frac{|\{e|e\in E, index(e)\leq k\}|}{k}
\end{equation*}
where $index(e)$ is the ranked index of $e$ which is predicted in $G$.

In classification, we use $micro-F1$ and $macro-F1$ to measure the performance of algorithms. For a label $L$, $TP(L)$, $FP(L)$ and $FN(L)$ are the number of true positives, false positives and false negatives in the instances which are predicted as $L$, $C$ is the label set:
\begin{equation*}
\begin{split}
&P(L) = \frac{TP(L)}{TP(L)+FP(L)}\\
&R(L) = \frac{TP(L)}{TP(L)+FN(L)}\\
&P(C) = \frac{\sum_{L\in C}TP(L)}{\sum_{L\in C}(TP(L)+FP(L))}\\
&R(C) = \frac{\sum_{L\in C}TP(L)}{\sum_{L\in C}(TP(L)+FN(L))}\\
&micro-F1 = \frac{2\times P(C)\times R(C)}{P(C)+R(C)}\\
&macro-F1 = \frac{\sum_{L\in C}\frac{2\times P(L)\times R(L)}{P(L)+R(L)}}{|C|}
\end{split}
\end{equation*}

Considering that each dataset has 14 snapshots in it, and the algorithms will be applied on each snapshot, so we choose the average performance as the final result.

\section{Results and Analysis}
\label{sec:result}
\subsection{Reconstruction}
\label{sec:reconstruction}
\begin{figure*}[htbp]
\centering
\subfigure[CA-Condmat]{
\begin{minipage}[t]{0.45\linewidth}
\centering
\includegraphics[width=\linewidth]{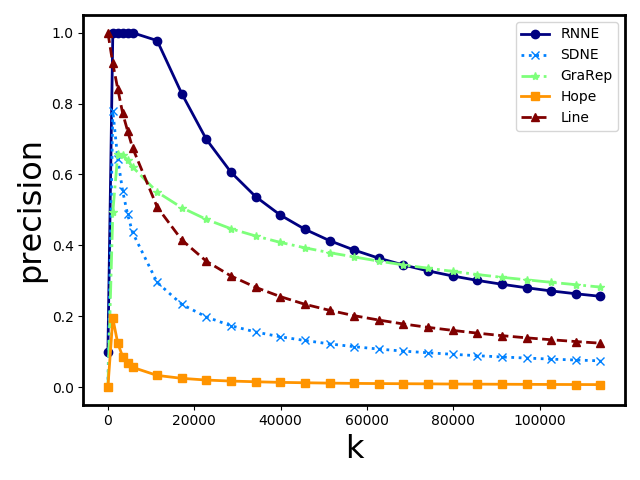}
\end{minipage}
}
\subfigure[CA-HepPh]{
\begin{minipage}[t]{0.45\linewidth}
\centering
\includegraphics[width=\linewidth]{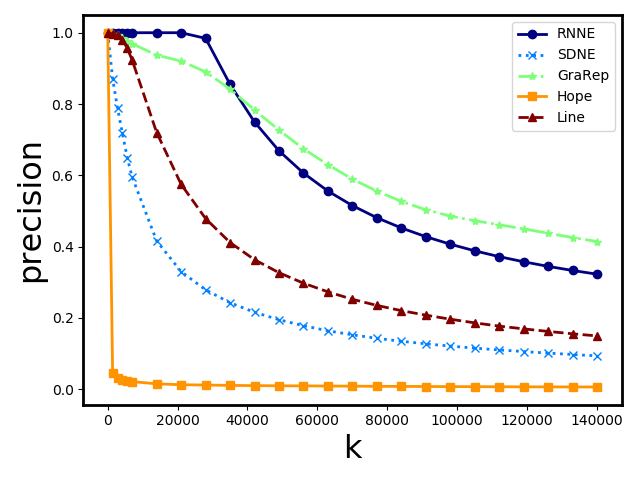}
\end{minipage}
}
\caption{\textbf{average \bm{$precision@k$} on CA-Condmat and CA-HepPh in reconstruction.}}
\label{fig:reconstruction}
\end{figure*}

Reconstruction means restituting the original network information from the embedding. In this task, we calculate the distance of each pair of nodes in the embedding space to measure the first-order proximity of the nodes. And then infer the edges using the proximity calculated above. There are the results of dataset CA-Condmat and CA-HepPh in Figure~\ref{fig:reconstruction}.

From this result, we can see RNNE archives better than SDNE, Hope and Line on these two datasets. And when $k$ is not big, RNNE also do better than GraRep. The algorithms who consider the first-order proximity or high-order proximity (RNNE, SDNE, GrapRep, Line) obviously perform better than those who doesn't (Hope). This result show that high-order proximity is very helpful to  preserve the original network structure. In fact, the history network information RNNE used is actually a noise in the reconstruction of current network. So RNNE has disadvantage on network  reconstruction theoretically.

\subsection{Classification}
\label{sec:classification}
\begin{figure*}[htbp]
\centering
\subfigure[blogCatalog]{
\begin{minipage}[t]{0.31\linewidth}
\centering
\includegraphics[width=\linewidth]{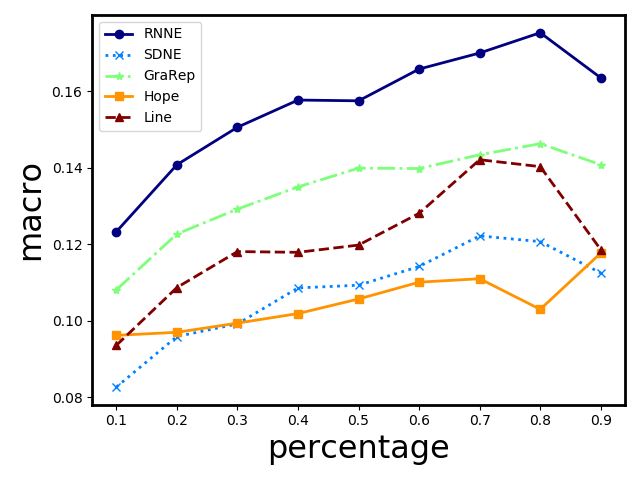}
\includegraphics[width=\linewidth]{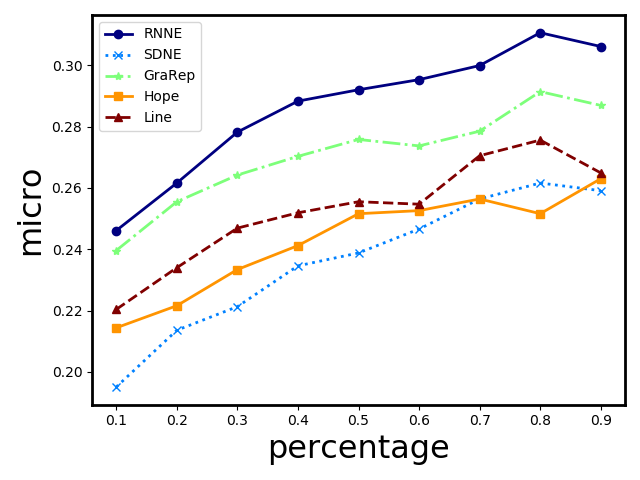}
\end{minipage}
}
\subfigure[Wiki]{
\begin{minipage}[t]{0.31\linewidth}
\centering
\includegraphics[width=\linewidth]{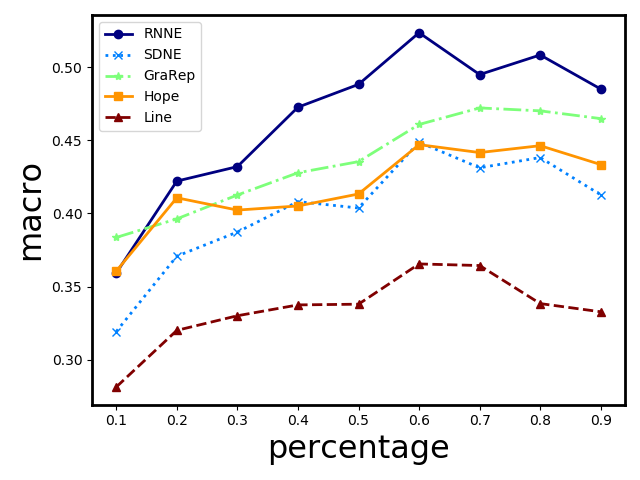}
\includegraphics[width=\linewidth]{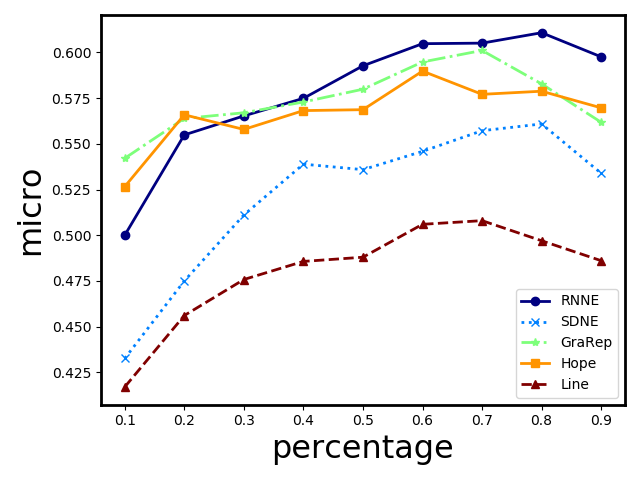}
\end{minipage}
}
\subfigure[email-Eu-core]{
\begin{minipage}[t]{0.31\linewidth}
\centering
\includegraphics[width=\linewidth]{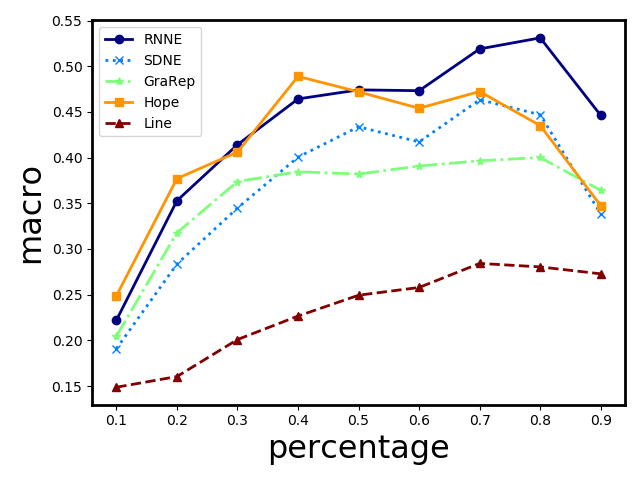}
\includegraphics[width=\linewidth]{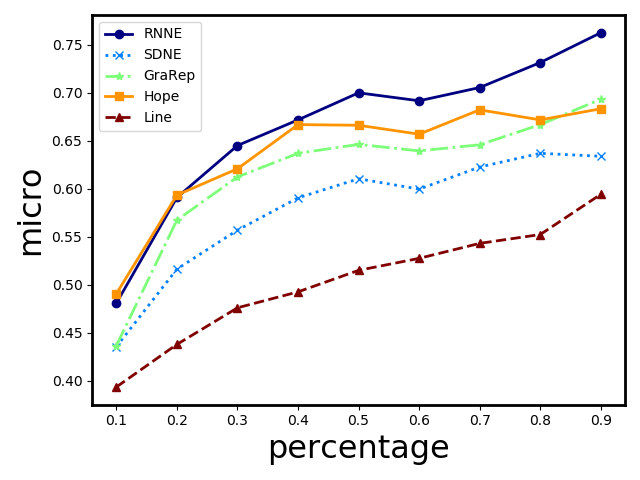}
\end{minipage}
}
\caption{\textbf{Micro-F1 and Macro-F1 on three datasets when the training percentage change.}}
\label{fig:classification}
\end{figure*}

Classification is a very common and important task in daily research and work. In this experiment, we use the node embedding as feature to classify each node into a label and then compare with its ground truth. Specifically, we use the LIBLINEAR~\cite{LIBLINEAR} as the solver of the classifiers. When training the classifiers, we randomly choose a part of nodes and their labels to train and use the rest to test. For Wiki, blogCatalog and email-Eu-core, we randomly choose 10\% to 90\% nodes to train. The results are shown in Figure~\ref{fig:classification}.

From this result, we can see RNNE make better performance than other four algorithms generally. The autoencoder structure of RNNE model can possibly make the nodes who are close in the feature space still be close in the embedding space. And when calculating node feature, we use the high-order proximity which are  more expressive than adjacency matrix.

\subsection{Link Predictions}
\label{sec:Link Prediction}
\begin{figure*}[htbp]
\centering
\subfigure[CA-Condmat]{
\begin{minipage}[t]{0.45\linewidth}
\centering
\includegraphics[width=\linewidth]{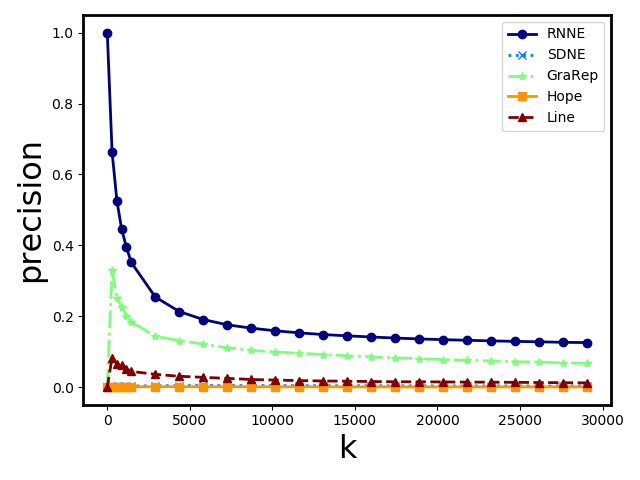}
\end{minipage}
}
\subfigure[CA-HepPh]{
\begin{minipage}[t]{0.45\linewidth}
\centering
\includegraphics[width=\linewidth]{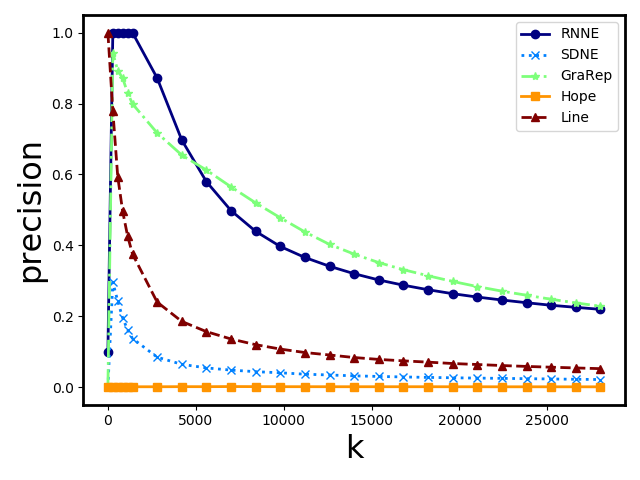}
\end{minipage}
}
\caption{\textbf{average \bm{$precision@k$} on CA-Condmat and CA-HepPh in link predictions.}}
\label{fig:link prediction}
\end{figure*}

Link predictions is a little similar with reconstruction, because they both need to judge whether an edge exist. Before doing this experiment, we will first randomly hide 15\% edges in the test networks, and then using their embedding to predict the hidden edges. To pay attention, when calculate $precision@k$, we will ignore the edges who are predicted but already exist in the after-hidden network. There are the results of dataset CA-Condmat and CA-HepPh in Figure~\ref{fig:link prediction}.

When $k$ become larger and larger, in the beginning, RNNE gets higher $precision@k$ than others, and afterwards, GraRep may do a little better. At most of the time in real world tasks such as recommendation, it doesn't require to predict too many links. On the one hand, with the predicting goes on, it will inevitably reduce the accuracy. On the other hand, people pay more attention to the pair of nodes who are most likely have a link. So it's very important to get higher precision when $k$ is small.

\subsection{Parameter Influence}
\label{sec:Parameter Influence}
\begin{figure*}[h]
\centering
\subfigure[Influence of $\alpha$ ($\beta$=5, $\gamma$=5)]{
\begin{minipage}[t]{\linewidth}
\centering
\includegraphics[width=0.32\linewidth]{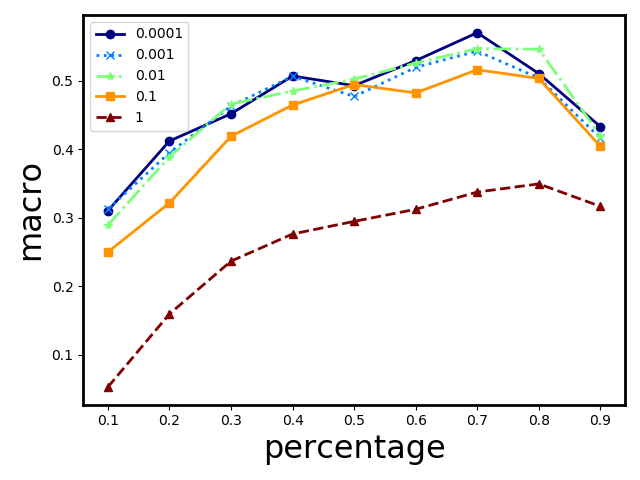}
\includegraphics[width=0.32\linewidth]{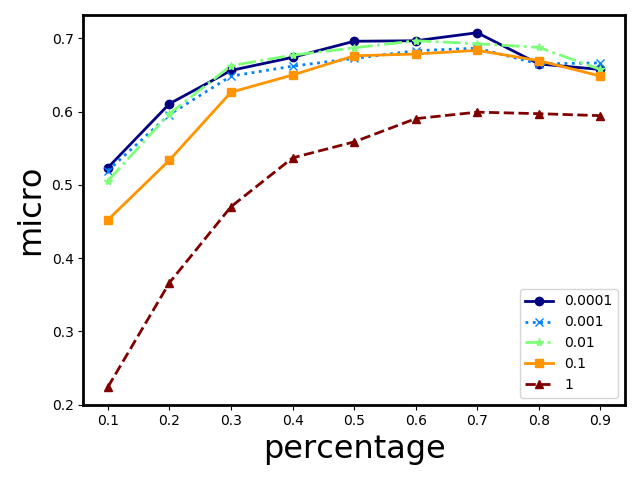}
\includegraphics[width=0.32\linewidth]{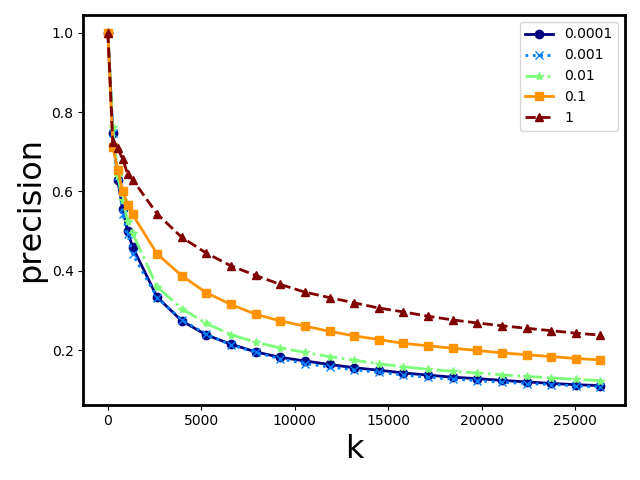}
\end{minipage}
}
\subfigure[Influence of $\beta$ ($\alpha$=0.1, $\gamma$=0)]{
\begin{minipage}[t]{\linewidth}
\centering
\includegraphics[width=0.32\linewidth]{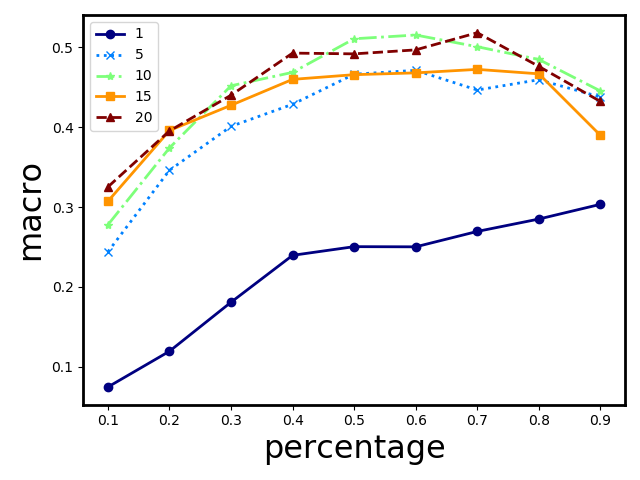}
\includegraphics[width=0.32\linewidth]{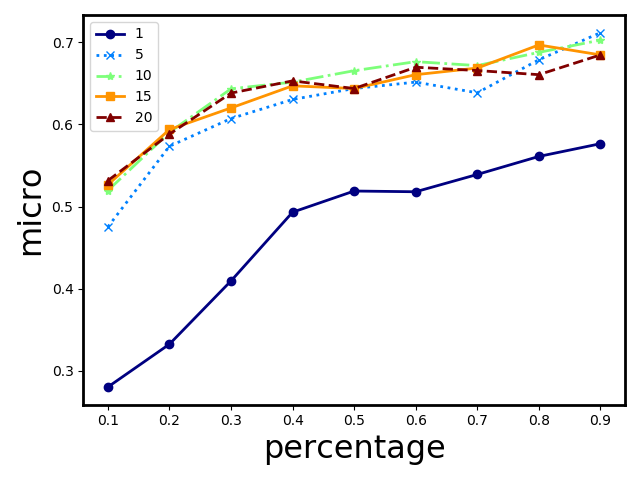}
\includegraphics[width=0.32\linewidth]{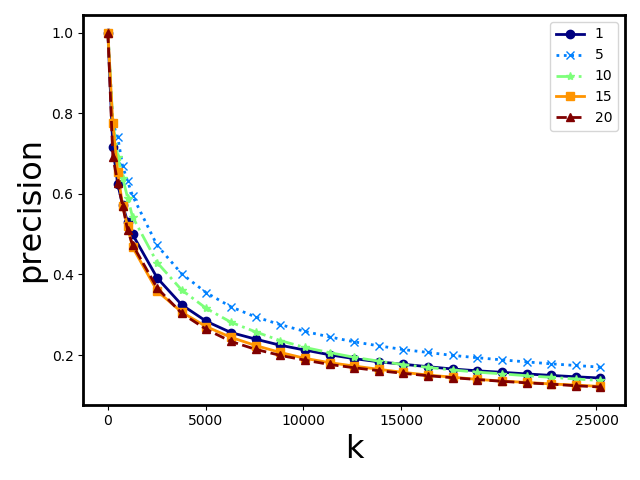}
\end{minipage}
}
\subfigure[Influence of $\gamma$ ($\alpha$=0.001, $\beta$=5)]{
\begin{minipage}[t]{\linewidth}
\centering
\includegraphics[width=0.32\linewidth]{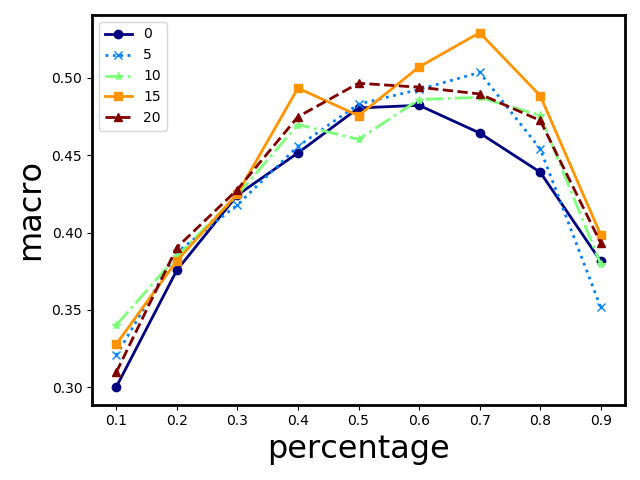}
\includegraphics[width=0.32\linewidth]{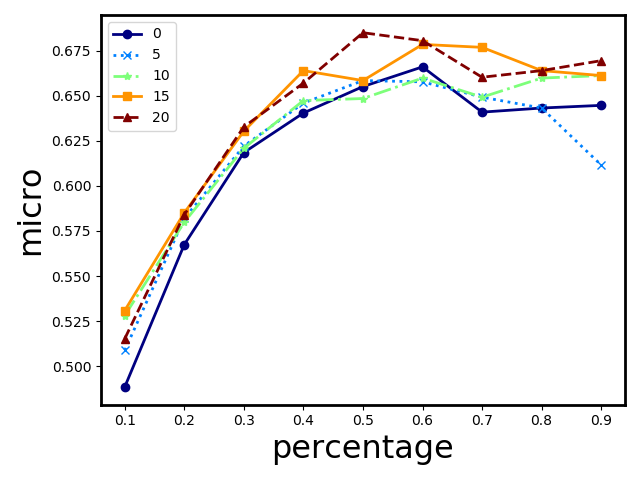}
\includegraphics[width=0.32\linewidth]{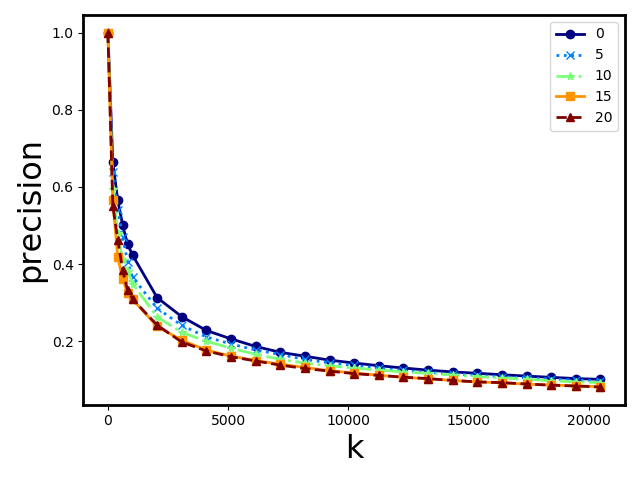}
\end{minipage}
}
\caption{\textbf{The influence of  parameters \bm{$\alpha$}, \bm{$\beta$} and \bm{$\gamma$} on the dataset of email-Eu-core.}}
\label{fig:Parameter Influence}
\end{figure*}

In this section, we  investigate the parameter influence to prove they are really effective for our tasks. Specifically, we evaluate the parameters $\alpha$, $\beta$ and $\gamma$ on the dataset of email-Eu-core. The results are shown in Figure~\ref{fig:Parameter Influence}.

In Figure~\ref{fig:Parameter Influence}(a), we can see the performance in classification and reconstruction when $\alpha$ varies in $[10^{-4}, 1]$ and  $\beta$ = 5, $\gamma$ = 5. It is very obvious when $\alpha$ becomes larger, the $precision@k$ is higher. But when $\alpha=1$, the quantity of classification become significantly worse. $\alpha$ is the weight of first-order proximity in the loss function, so the larger $\alpha$ is, the more the model is concerned on the direct links between nodes. It is important to find a balance between first-order and high-order proximity.

In Figure~\ref{fig:Parameter Influence}(b), we can see the performance when $\beta$ varies in [1, 20] and $\alpha$ = 0.1, $\gamma$ = 0. $\beta$ is the weight of non-zero part when reconstructing the node feature in the autoencoder. When $\beta$ = 1, which means the non-zero elements and zero elements have the same weight, the results are not good. However, when $\beta$ is too large, the $precison@k$ in reconstruction task is still not good enough (in this experiment, $\beta$ = 5 is the best) since too large $\beta$ makes the autoencoder ignore the information in zero element. Thus, we should pay more attention to the non-zero elements and still concentrate on zero elements properly.

In Figure~\ref{fig:Parameter Influence}(c), we can see the performance when $\gamma$ varies in [0, 20] and $\alpha$ = 0.001, $\beta$ = 5. $\gamma$ is used to reduce the difference of the same node representations at different time. That means, the embedding results not only depend on the current network, but also depend on the previous. So we can see when $\gamma$=0, the $precision@k$ is the best, though it still has ``noise'' because of the RNN structure. Of course, the larger $\gamma$ gain better performance than $\gamma$ = 0 in classification. So the choose of $\gamma$ is depends on whether we focus more on network structure or node feature.

\section{Conclusion}
\label{sec:Conclusion}
In this paper, we propose Recurrent Neural Network Embedding (RNNE), an algorithm for dynamic network embedding with deep neural network. In order to unify the input network structure at different time, we add virtual nodes and replace virtual nodes with real nodes when nodes changing happened. In the method of embedding,  RNNE not only keeps the local and global network structure via first-order and high-order proximity, but also reduces the influence of noise by transferring the previous embedding information. We compare RNNE with several other algorithms on various datasets and tasks, and then show the parameters influence on the performance of embedding. The results show that our method is effective and can achieve better performance than other algorithms on the tested datasets. 

The future work will try to use the probabilistic models \cite{chen-2009-pcvm,chen-2014-epcvm}, its multi-objective version \cite{lyu2019multiclass} and large-scale version \cite{jiang2017scalable} to incorporate with dynamic network embedding. In addition, the ensemble methods \cite{chen2009regularized,chen2009predictive,chen2010multiobjective} could be employed to improve the performance of embedding and the following applications. 

\bibliographystyle{unsrt} 
\bibliography{sample-base}  


\end{document}